\documentclass[11pt,letterpaper]{article}
\pdfoutput=1
\usepackage{arabtex}
\usepackage{emnlp2016}

\usepackage{times}
\usepackage{latexsym}
\usepackage[utf8]{inputenc}

\usepackage{algorithm}
\usepackage{algorithmic}
\usepackage{amsmath}
\usepackage{amsthm}
\usepackage{amssymb}
\usepackage{graphicx}
\usepackage{xspace}
\usepackage{tabularx}
\usepackage{multicol}
\usepackage{multirow}
\usepackage{url}
\usepackage{wrapfig}
\usepackage{comment}
\usepackage{listings}
\usepackage[utf8]{inputenc}
\usepackage{fancyvrb}
\usepackage{booktabs}
\usepackage{color,soul}
\usepackage[normalem]{ulem}

\newcommand{\bmx}[0]{\begin{bmatrix}}
\newcommand{\emx}[0]{\end{bmatrix}}

\newcommand{\vect}[1]{\mathbf{#1}}

\newcommand{\vc}[0]{\vect{c}}

\newcommand{\vh}[0]{\vect{h}}

\newcommand{\vz}[0]{\vect{z}}

\emnlpfinalcopy

\def\emnlppaperid{***}

\title{First Result on Arabic Neural Machine Translation}

\author{Amjad Almahairi \\ MILA, Universit\'e de Montr\'eal \\ 
  amjad.almahairi@umontreal.ca
\And Kyunghyun Cho  \\ New York University \\ kyunghyun.cho@nyu.edu \\
\AND Nizar Habash \\ New York University \\ nizar.habash@nyu.edu
\And Aaron Courville\\ MILA, Universit\'e de Montr\'eal \\ aaron.courville@umontreal.ca}

\date{}

\begin{document}

\maketitle

\setarab
\novocalize

\begin{abstract}
    Neural machine translation has become a major alternative to widely used
    phrase-based statistical machine translation. We notice however that much of
    research on neural machine translation has focused on European languages
    despite its language agnostic nature. In this paper, we apply neural machine
    translation to the task of Arabic translation
    (Ar$\leftrightarrow$En) and compare it against a standard phrase-based
    translation system. We run extensive comparison using various configurations
    in preprocessing Arabic script and show that the phrase-based and neural
    translation systems perform comparably to each other and that proper
    preprocessing of Arabic script has a similar effect on both of the systems.
    We however observe that the neural machine translation significantly
    outperform the phrase-based system on an out-of-domain test set, making it
    attractive for real-world deployment. 
\end{abstract}

\section{Introduction}

Neural machine
translation~\cite{kalchbrenner2013recurrent,sutskever2014sequence,cho2014learning}
has become a major alternative to the widely used statistical phrase-based
translation system~\cite{koehn2003statistical}, evidenced by the successful
entries in WMT'15 and WMT'16.


Previous work on using neural networks for Arabic translation has mainly focused
on using neural networks to induce an additional feature for phrase-based
statistical machine translation systems (see, e.g.,
\cite{devlin2014fast,setiawan2015statistical}). This hybrid approach has
resulted in impressive improvement over other systems without any neural
network, which raises a hope that a fully neural translation system may achieve
a even higher translation quality.  We however found no prior work on applying a
fully neural translation system (i.e., neural machine translation) to 
Arabic translation.

In this paper, our aim is therefore to present the first result on the Arabic
translation using neural machine translation. On both directions (Ar$\to$En and
En$\to$Ar), we extensively compare a vanilla attention-based neural machine
translation system~\cite{bahdanau2014neural} against a vanilla phrase-based
system (Moses, \cite{koehn2003statistical}), while varying pre-/post-processing routines, including morphology-aware tokenization and
orthographic normalization, which were found to be crucial in
Arabic translation
\cite{habash2006arabic,badr2008segmentation,el2012orthographic}. 

The experiment reveals that neural machine translation performs comparably to
the standard phrase-based system. We further observe that
the tokenization and normalization routines, initially proposed for phrase-based
systems, equally improve the translation quality of neural machine translation.
Finally, on the En$\to$Ar task, we find the neural translation system to be more
robust to the domain shift compared to the phrase-based system.

\section{Neural Machine Translation}
\label{sec:nmt}

A major workforce behind neural machine translation is an
attention-based encoder-decoder
model~\cite{bahdanau2014neural,cho2015describing}. This attention-based
encoder-decoder model consists of an encoder, decoder and attention mechanism.
The encoder, which is often implemented as a bidirectional recurrent
network, reads a source sentence $X=(x_1, \ldots, x_{T_x})$
and returns a set of context vectors $C=(\vh_1, \ldots, \vh_{T_x})$.

The decoder is a recurrent language model. At each time $t'$, it
computes the new hidden state by
\[
    \vz_{t'} = \phi(\vz_{t'-1}, \tilde{y}_{t'-1}, \vc_{t'}),
\]
where $\phi$ is a recurrent activation function, and $\vz_{t'-1}$ and
$\tilde{y}_{t'-1}$ are the previous hidden state and previously decoded target
word respectively. $\vc_{t'}$ is a time-dependent context vector and is a
weighted sum of the context vectors returned by the encoder:
$
    \vc_{t'} = \sum_{t=1}^{T_x} \alpha_t \vh_t,
$
where the attention weight $\alpha_t$ is computed by the attention mechanism
$f_{\text{att}}$:
\mbox{
$
    \alpha_t \propto \exp(f_{\text{att}}(\vz_{t'-1}, \tilde{y}_{t'-1},
    \vh_{t})).
$
}
In this paper, we use a feedforward network with a single $\tanh$ hidden layers
to implement $f_{\text{att}}$.

Given a new decoder state $\vz_{t'}$, the conditional distribution over the next
target symbol is computed as
\begin{align*}
    p(y_t = w| \tilde{y}_{<t}, X) \propto \exp(g_w(\vz_{t'})),
\end{align*}
where $g_w$ returns a score for the word $w$, and $V$ is a target vocabulary.

The entire model, including the encoder, decoder and attention mechanism, is
jointly tuned to maximize the conditional log-probability of a ground-truth
translation given a source sentence using a training corpus of parallel sentence
pairs. This learning process is efficiently done by stochastic gradient descent
with backpropagation. 

\paragraph{Subword Symbols}

\newcite{sennrich2015neural}, \newcite{chung2016character} and
\newcite{luong2016achieving} showed that the attention-based neural translation
model can perform well when source and target sentences are represented as
sequences of subword symbols such as characters or frequent character $n$-grams.
This use of subword symbols elegantly addresses the issue of large target
vocabulary in neural networks \cite{jean2014using}, and has become a {\it de
facto} standard in neural machine translation.
Therefore, in our experiments, we use character $n$-grams selected by byte pair
encoding \cite{sennrich2015neural}.

\section{Processing of Arabic for Translation}

\subsection{Characteristics of Arabic Language}
\label{sec:arabic}

Arabic exhibits a rich morphology. This makes Arabic challenging for
natural language processing and machine translation. For instance, a
single Arabic token
`\RL{wlmrkbth}' (`and to his vehicle' in English)
is formed by prepending `\RL{w}' (`and') and `\RL{l--}' (`to') to the base
lexeme `\RL{mrkbT}' (`vehicle'), appending `\RL{h}' (`his') and
replacing the feminine suffix `\RL{T}' ({\it ta marbuta}) of the base
lexeme to `\RL{t}'. This feature of Arabic is challenging, as (1) it
increases the number of out-of-vocabulary tokens, (2) it consequently
worsens the issue of data sparsity \footnote{see Sec.~5.2.1 of
  \cite{cho2015natural} for detailed discussion.}, and (3) it
complicates the word-level correspondence between Arabic and another
language in translation. This is often worsened by the orthographic
ambiguity found in Arabic scripts, such as the inconsistency in
spelling certain letters.


Previous work has thus proposed morphology-aware tokenization and
orthographic normalization as two crucial components for building a
high quality phrase-based machine translation system (or its variants)
for
Arabic~\cite{habash2006arabic,badr2008segmentation,el2012orthographic}. These
techniques have been found very effective in alleviating the issue of
data sparsity and improving the generalization to tokens not included
in a training corpus (in their original forms.)

\subsection{Morphology-Aware Tokenization}

The goal of morphology-aware tokenization, or morpheme
segmentation~\cite{creutz2005unsupervised} is to split a word in its surface
form into a sequence of linguistically sound sub-units. Contrary to simple
string-based tokenization methods, morphology-aware tokenization relies on
linguistic knowledge of a target language (Arabic in our case) and applies, for
instance, various morphological or orthographic adjustments to the resulting
sub-units. 

In this paper, we investigate the tokenization scheme used in the Penn Arabic
Treebank (ATB, \cite{maamouri2004penn})
which was found to work well with
phrase-based translation system in \cite{el2012orthographic}. This
tokenization separates all clitics other than definite articles. 

When translating {\it to} Arabic, the decoded sequence of tokenized symbols must be
{\it de-tokenized}. This de-tokenization step is not trivial, as it needs to
undo any adjustment (implicitly) made by the tokenization algorithm. In this
work, we follow the approach proposed
in~\cite{badr2008segmentation,salameh2015matters}. This approach builds a
lookup table from a training corpus and uses it for mapping a tokenized form
back to its original form. When the tokenized form is missing in the
lookup table, we back off to a number of hand-crafted de-tokenization rules.

\subsection{Orthographic Normalization}

Since the sources of major orthographic
ambiguity are in the letters `alif' and `ya', we normalize
these letters (and their inconsistent replacements.) Furthermore, we replace
parentheses `(' and `)' with special tokens `--LRB--' and `--RRB--',
and remove diacritics.

\section{Experimental Settings}


\subsection{Data Preparation}

\paragraph{Training Corpus}
We combine LDC2004T18, LDC2004T17 and LDC2007T08 to form a training parallel
corpus. The combined corpus contains approximately 1.2M sentence pairs, with 33m
tokens on the Arabic side. Most of the sentences are from news articles. 
We ignore sentence pairs which either side has more
than 100 tokens.

\paragraph{In-Domain Evaluation Sets}
We use the evaluation sets from NIST 2004 (MT04) and 2005 (MT05) as development
and test sets respectively. In Ar$\to$En, we use all four English reference
translations to measure the translation quality. We use only the first English
sentence out of four as a source during En$\to$Ar. Both of these sets are
derived from news articles, just as the training corpus is. 

\paragraph{Out-of-Domain Evaluation Set}
In the case of En$\to$Ar, we evaluate both phrase-based and neural translation
systems on MEDAR evaluation set~\cite{hamon2011evaluation}. This set has four
Arabic references per English sentence. It is derived from web pages discussing
climate changes, significantly differing from the training corpus. This set was
selected to highlight the robustness to domain mismatch between training and
test sets. 

We verify the domain mismatches of the evaluation sets relative to the training
corpus by fitting a 5-gram language model on the training corpus and computing
the likelihoods of the evaluation sets, on the Arabic side. As can be seen in
Table~\ref{table:lm-scores}, the domain of the MEDAR is significantly further away
from the training corpus than the others are. 

\begin{table}[t]
  \centering 
  \small
  \begin{tabular}{c||cc| c}
    & MT04   & MT05   & MEDAR    \\
    \toprule
    Avg. Log-Prob. & -59.74 & -55.97 & -75.03   
  \end{tabular}
  \caption{Language model scores of the Arabic side.
  The language model was tuned on the training corpus.}
  \label{table:lm-scores}

  \vspace{-6mm}
\end{table}

\paragraph{Note on MT04 and MT05} 
We noticed that a significant portion of Arabic
sentences in MT04 and MT05 are found verbatim in the training corpus (172 on
MT04 and 26 on MT05). 
In order to accurately measure the generalization performance, we
removed those duplicates from the evaluation sets.

\begin{table*}[t]
  \centering \small

    \begin{tabular}{c | c c c | c c || c c | c c || c c}      
      & \multicolumn{3}{c|}{Arabic} & \multicolumn{2}{c||}{English}  & \multicolumn{4}{c||}{En$\to$Ar} & \multicolumn{2}{c}{Ar$\to$En} \\
      & Tok. & Norm. & ATB & Tok. & Lower    & \multicolumn{2}{c|}{MT05} & \multicolumn{2}{c||}{MEDAR} & \multicolumn{2}{c}{MT05}\\
      \toprule
      \multirow{6}{*}{\rotatebox[origin=c]{90}{PB-SMT}} & $\surd$ & & & $\surd$ & & 31.52 & -- &  8.69 & -- & 48.59 & -- \\
      &$\surd$ & & & $\surd$ & $\surd$ & 33.03 & (1.51) & 9.78 & (1.09)  & 49.44 & (0.85) \\
      &$\surd$ & $\surd$ &   & $\surd$ & & 34.98 & (3.46) & 17.34 & (8.65) & 49.51 & (0.92)\\
      &$\surd$ & $\surd$ &   & $\surd$ & $\surd$ & 35.63 & (4.11) & 17.75 & (9.06) & 49.91 & (1.32)\\
      &$\surd$ & $\surd$ & $\surd$ & $\surd$ & & 35.7 & (4.18) & 18.67 & (9.98) & 50.67 & (2.08)\\
      &$\surd$ & $\surd$ & $\surd$ & $\surd$ & $\surd$ & 35.98 & (4.46) & 19.27 & (10.58)  & 51.19 & (2.60) \\
      \midrule
      \multirow{6}{*}{\rotatebox[origin=c]{90}{Neural MT}} & $\surd$ & & & $\surd$ & & 28.64 & -- & 11.09 & --  & 47.12 & --  \\
      &$\surd$ & & & $\surd$ & $\surd$ & 29.77 & (1.13) & 10.15 & (-0.94)  & 47.63 & (0.51) \\
      &$\surd$ & $\surd$ & & $\surd$ & & 32.53 & (3.89) & 22.36 & (11.27)  & 48.53 & (1.41)\\
      &$\surd$ & $\surd$ & & $\surd$ & $\surd$ & 32.95 & (4.31) & 22.79 & (11.70) & 47.53 & (0.41)\\
      &$\surd$ & $\surd$ & $\surd$ & $\surd$ & & 33.53 & (4.89) & 23.11 & (12.02) & 49.21 & (2.09) \\
      &$\surd$ & $\surd$ & $\surd$ & $\surd$ & $\surd$ & 33.62 & (4.98) & 24.46 & (13.37) & 49.7  & (2.58)\\
    \end{tabular}
    \caption{BLEU scores with the improvement over the tokenization-only models
        in the parentheses.}
    \label{table:result-en-ar}
    \vspace{-4mm}
\end{table*}

\subsection{Machine Translation Systems}

\paragraph{Phrase-based Machine Translation}
We use Moses~\cite{koehn2007moses} to build a standard phrase-based statistical
machine translation system.
Word alignment was extracted by
GIZA++~\cite{och2003systematic}, and we used phrases up to 8 words to build a
phrase table. We use the following options for alignment symmetrization
and reordering model: {\it grow-diag-final-and} and {\it msd-bidirectional-fe}.
KenLM~\cite{Heafield-estimate} is used as a language model and trained on the target side
of the training corpus. 

\paragraph{Neural machine translation}
We use a publicly available implementation of attention-based neural machine
translation.\footnote{
    \scriptsize \url{https://github.com/nyu-dl/dl4mt-tutorial}
} For both directions--En$\to$Ar and Ar$\to$En--, the encoder is a bidirectional
recurrent network with two layers of 512$\times$2 gated recurrent units
(GRU,~\cite{cho2014learning}), and the decoder a unidirectional recurrent
network with 512 GRU's. Each model is trained for approximately seven days using
Adadelta~\cite{zeiler2012adadelta} until the cost on the development set stops
improving. We regularize each model by applying
dropout~\cite{srivastava2014dropout} to the output layer and penalizing the L2
norm of the parameters (coefficient $10^{-4}$). We use beam search with width
set to 12 for decoding.

\subsection{Normalization and Tokenization}

\paragraph{Arabic}
We test {\it simple tokenization} ({\bf Tok}) based on the script from
Moses, and orthographic {\it normalization} ({\bf Norm}), and {\it
  morphology-aware tokenization} ({\bf ATB}) using
MADAMIRA~\cite{pasha2014madamira}, . In the latter scenario, we
reverse the tokenization before computing BLEU.  Note that
\textbf{ATB} includes \textbf{Norm}, and both of them include simple
tokenization performed by MADAMIRA.

\paragraph{English}
We test {\it simple tokenization} ({\bf Tok}), lowercasing ({\bf Lower}) for
En$\to$Ar and {\it truecasing} ({\bf True}, \cite{lita2003truecasing}) for
Ar$\to$En. 

\paragraph{Byte pair encoding}
As mentioned earlier in Sec.~\ref{sec:nmt}, we use byte pair encoding
(BPE) for neural machine translation. We apply BPE to the
already-tokenized training corpus to extract a vocabulary of up to 20k
subword symbols. We use the publicly available script released by
\newcite{sennrich2015neural}.

%

\section{Result and Analysis}

\noindent{\bf En$\to$Ar}
From Table~\ref{table:result-en-ar}, we observe that the translation quality
improves as a better preprocessing routine is used. By using the normalization
as well as morphology-aware tokenization (Tok+Norm+ATB), the phrase-based and
neural systems each achieve as much as +4.46 and +4.98 BLEU over the baselines,
on MT05. The improvement is even more apparent on the MEDAR whose domain
deviates from the training corpus, confirming that proper
preprocessing of Arabic script indeed helps in handling
word tokens that are not present in a training corpus.

We notice that the tested tokenization strategies have nearly identical effect
on both the phrase-based and neural translation systems. The translation quality
of either system is mostly effective by the tokenization strategy employed for
Arabic, and is largely insensitive to whether source sentences in English were
lowercased. This reflects well the complexity of Arabic scripts, compared to
English, discussed earlier in Sec.~\ref{sec:arabic}.

Another important observation is that the neural translation system
significantly outperforms the phrase-based one on the out-of-domain evaluation
set (MEDAR), while they perform comparably to each other in the case of the
in-domain evaluation set (MT05). We conjecture that this is due to the neural
translation system's superior generalization capability based on its use of
continuous space representations.

\noindent{\bf Ar$\to$En} 
In the last column of Table~\ref{table:result-en-ar}, we observe a trend similar
to that from En$\to$Ar. First, both phrase-based and neural machine translation
benefit quite significantly from properly normalizing and tokenizing Arabic,
while they are both less sensitive to truecasing English. The best translation
quality using either model was achieved when all the tokenization methods were
applied (Ar: Tok+Norm+ATB and En:Tok+True), improving upon the baseline by more
than 2+ BLEU.  Furthermore, we again observe that the phrase-based and neural
translation systems perform comparably to each other.

\section{Conclusion}
We have provided first results on Arabic neural MT, and performed
extensive experiments comparing it with a standard phrase-based
system. We have concluded that neural MT benefits from morphology-based
tokenization and is robust to domain change.

\bibliography{emnlp2016} \bibliographystyle{emnlp2016}

\end{document}